\documentclass[12pt]{article}
\usepackage{times}
\usepackage[margin=1in]{geometry}
\usepackage[doublespacing]{setspace}
\usepackage{natbib}

\usepackage{times}
\usepackage{latexsym}

\usepackage[T1]{fontenc}

\usepackage[utf8]{inputenc}

\usepackage{microtype}

\usepackage{inconsolata}

\usepackage{graphicx} 
\usepackage{subfigure} 
\usepackage{color} 
\usepackage{arydshln} 
\usepackage{amsmath, amsfonts, amssymb, bm} 
\usepackage{float} 
\usepackage{placeins} 
\usepackage{multirow} 
\usepackage{xcolor} 
\usepackage{enumitem} 
\usepackage{tablefootnote} 
\usepackage{longtable}
\usepackage[normalem]{ulem} 
\usepackage{algorithm, algorithmic} 
\usepackage{hyperref}

\DeclareMathAlphabet{\mathbbmsl}{U}{bbm}{m}{sl}

\definecolor{lightred2}{HTML}{ea9999}
\definecolor{lightyellow2}{HTML}{ffe599}
\definecolor{lightblue2}{HTML}{9fc5e8}
\definecolor{lightgreen2}{HTML}{b6d7a8}

\definecolor{lightred1}{HTML}{e06666}
\definecolor{lightyellow1}{HTML}{ffd966}
\definecolor{lightblue1}{HTML}{6fa8dc}
\definecolor{lightgreen1}{HTML}{93c47d}

\usepackage{xspace}

   \usepackage[disable]{todonotes} 
\newcommand{\ignore}[1]{\textcolor{gray}{}}

\newcommand{\note}[4][]{\todo[author=#2,color=#3,size=\scriptsize,fancyline,#1]{#4}} 

\newcommand{\huda}[2][]{\note[#1]{Huda}{violet!20}{#2}}

\newcommand{\esl}{ESL\xspace}

\def\Snospace~{\S{}}

\usepackage{microtype}

\usepackage{ tipa }

\author{Regular Paper}
\date{}

\title{How to Choose How to Choose Your Chatbot:\\ A  \textit{M}assively \textit{M}ulti-\textit{S}ystem \textit{M}ulti\textit{R}eference Data Set \\ for Dialog Metric Evaluation}

\begin{document}

\maketitle








\begin{abstract}

We release MMSMR,\footnote{Pronounced like \textbf{mesmer}ize.} a \textbf{M}assively \textbf{M}ulti-\textbf{S}ystem \textbf{M}ulti\textbf{R}eference dataset to enable future work on dialog metrics and evaluation.
Automatic metrics for dialogue evaluation should be robust proxies for human judgments; however, the verification of robustness is currently far from satisfactory. To quantify the robustness correlation and understand what is necessary in a test set, we create and release an 8-reference dialog dataset by extending single-reference evaluation sets and introduce a  new language learning conversation dataset.
We then train 1750 systems and evaluate them and publicly available large models on our novel test set and the DailyDialog dataset. In addition to the novel test set, we release  model hyper parameters, inference outputs, and metric scores for each system on a variety of datasets. 

\end{abstract}

\ignore{
contributions (dataset?)
 - multi-turn dataset w/ multiple references (9+) - ESL 2/3 (1k) + NCM (200) + DailyDialog (? 100) 
 - LOTS of models (same arch) (1400+) -- 15 diff arch
 - evaluation metrics (30+)
 - ??? human evaluations
     - what's the best systems _not_ how good are the metrics 
     - MT metrics track only on the strongest systems
     - any output that is the top X for any metric 
     
RQ:
 - examining parameters of the models: "what is the effect of parameters"
 - what is the correlation between different metrics? - does it change by quality
 - how important are multiple references? (DialogDialog baseline [Gupta] 100 12-way, $\Delta$BLEU paper)
       - annotation budget for X - 1 ref or better to have X/4 sentences w/ 4 references each
 - No current metric eval (high quality DSTC9 BAD) sets have $> 30$ models. Do we need lots of model diversity?
 - When do we see Simpson's Paradox -- if at all?
 -

 plots / results
  - Parallel coordinates plot
  - correlation plots (metrics)
  - correlations to human evaluations of models - replicate fig 1 Gupta et al.
  - BLEU (and all metrics) - same person (0.2) vs between people (0.6)

TBD:
 - human evals (this is relative to extreme valuates of metrics - for human evaluation - REALSumm did this.
 - inference ESL
 - run all metrics (1-2 weeks [Wall time])
 - 
 
 ** how do we use the checkpoints?
}

\section{Introduction}

Automatically evaluating social conversational agents (a.k.a. social dialogue systems or chatbots) is a challenging task that, if solved, would save time and money by making it easier to tune or evaluate such agents. There are three prevailing methods for evaluation: reference-based metrics $f(\hat{u}_t \mid \{r_t\})$, reference-free metrics $f(\hat{u}_t \mid u_{t-1} \dots, u_0)$, and perplexity $f(\hat{u}_t)$, where $\hat{u}_t$ is the model generated response, $\{r_t\}$ are a set of references, and $u_{t-1}$ is the previous utterance in the conversation. Evaluation metrics such as BLEU~\citep{papineni-etal-2002-bleu}, METEOR~\citep{banerjee-lavie-2005-meteor}, ROUGE~\citep{lin-2004-rouge}, BERTScore~\citep{zhang2019bertscore}, and BARTScore~\citep{yuan2021bartscore} are reported in the evaluation of open-domain chatbots models despite evidence of weak statistically significant correlation with human judgments \citep{liu-etal-2016-evaluate,yeh2021comprehensive,zhang2021automatic}. 
There is some evidence attributing the low correlation between reference-based metrics and human judgments to the ``one-to-many'' problem in conversational dialogue \citep{galley-etal-2015-deltableu,zhao-etal-2017-learning,gangal-etal-2021-improving}, whereby there can be multiple appropriate responses to a given input, and only a single `ground-truth' reference response is used.
Prior work demonstrated a higher correlation between automatic metrics and human judgments when using multiple references on the DailyDialog \citep{li-etal-2017-dailydialog} dataset \citep{gupta-etal-2019-investigating}. Building upon this work, we extend the investigation to other datasets and employ a distinct methodology for gathering human annotations. A limitation of prior datasets is that the number of systems evaluated is extremely sparse~\citep{zhang2021automatic}.

To address such limitations, we  (will, on publication)  release MMSMR,
a \textbf{M}assively \textbf{M}ulti-\textbf{S}ystem \textbf{M}ulti\textbf{R}eference dataset. 
\huda{MMSMR consists of a new conversational model evaluation dataset from a subset of the teaching English as a second language website (\href{http://www.rong-chang.com}{TESL}) which includes 1000 two and three-turn conversational prompts. We also generate multiple `ground truth' references for each prompt. Additionally, we collect multiple `ground-truth' responses for the one-turn handcrafted dataset (NCM) made by \citet{DBLP:journals/corr/VinyalsL15}.} 
Our contributions are:
\begin{itemize}[noitemsep,topsep=0pt]
    \item We create and release a new conversational evaluation dataset based on hand-crafted conversations from material for teaching English as a second language\footnote{\href{http://www.rong-chang.com}{\texttt{rong-chang.com}}} (\esl).\footnote{A subset of the prompts was made available online for use by other researchers in the past, but the dataset has not yet been published or released in full.} 
    \item We collect and release multiple diverse `ground-truth' human-generated reference responses for the \esl and NCM datasets.
    \item We train and release outputs of 1750 models on these data sets to enable the study of how metrics perform on a variety of models.
    \item We release the parameters to enable research on metrics without re-training new models.
    \item We release (sampled) inference from six large open-source models.
    \item We demonstrate the utility of the above contributions through analysis. 
\end{itemize}
MMSMR is designed to test the robustness of dialog evaluation metrics in a statistically robust way.
\section{Background \& Related Work}

Our work uses MMSMR to analyze automatic dialog metrics. We are far from the first to evaluate metrics using multiple annotations. Both multiple human-generate references, as well as multiple automatic references, have been explored \citep{gupta-etal-2019-investigating,galley-etal-2015-deltableu,gangal-etal-2021-improving}. In particular, \citeauthor{gangal-etal-2021-improving} demonstrate that  automatically expanded reference sets improve correlations between human ratings and automatic metrics. 

Other related prior work explores the relationships between metrics. In \citet{yeh2021comprehensive}, 23 automatic evaluation metrics are evaluated on 10 datasets which are assessed to compare their shortcomings and strengths. In contrast to our work, these datasets rarely contained multiple references and also had very few dialog systems. Similarly, \citet{deriu2021survey} surveys new evaluation methods that reduce human interaction.

While to the best of our knowledge large multi-system datasets do not exist for dialog evaluation, \citet{zhang-duh-2020-reproducible} did a grid search on machine translation
and released it for research in hyper parameter optimization.

\subsection{Metrics}
Automatic dialog evaluation metrics are mainly of two types: model based and rule based. The model based metrics measure the quality of responses that are generally trained. Rule-based metrics analyze the system response using heuristic rules based on human references and conversation context.

Several string overlap metrics are borrowed from other NLP tasks. In these metrics, the model output is compared to a human reference response. 
Bleu \cite{papineni-etal-2002-bleu}, and Meteor \cite{banerjee-lavie-2005-meteor} come from machine translation,
and Rouge \cite{lin-2004-rouge} comes from summarization. Bleu is based on string matches using n-gram precision of the responses Meteor includes synonyms and stems for computing the score. Rouge on the other hand uses n-gram recall.
The effectiveness of these word overlap metrics has been a source of great debate~\citep{liu-etal-2016-evaluate,lowe-etal-2017-towards,gupta-etal-2019-investigating,galley-etal-2015-deltableu}.

The first model based metrics 
compute similarity between context and reference word embeddings \cite{mikolov2013distributed,pennington2014glove,mikolov2013efficient}.
%
BERTScore \cite{zhang2019bertscore}  uses contextual embeddings for  computing token similarity.

Prism \cite{thompson-post-2020-automatic}  and BARTScore \cite{yuan2021bartscore} use sequence-level model scores. sequence-to-sequence paraphraser to score the output conditioned on human references, while BARTScore uses BART \cite{lewis-etal-2020-bart}, a denoising model. DialoRPT \cite{gao2020dialogue} is based on a set of GPT-2 models which are fine-tuned on a Reddit human feedback dataset.


USL-H \cite{phy2020deconstruct} is a metric that is flexible to a task where a method is proposed to compound metrics called USL-H, which is Understandability, Sensibleness, and Likability in Hierarchy which is a single metric. USL-H combines three different models valid utterance prediction (VUP), next sentence prediction (NSP), and masked language model (MLM) where each model is trained on different tasks. 


%
\section{Data Collection}
Here we describe our methods for collecting 3500 new multiturn conversations, collecting multiple references for each multiturn dataset, and collecting ratings for model generated responses. 

\subsection{Reference collection}
We created a HIT (human intelligence task) for Amazon's Mechanical Turk (AMT) to collect multiple references. Each worker was shown 10 one-, two-, or three-turn conversations and asked to provide 2 to 5 responses to the last turn in each conversation.\footnote{The HIT html is available in the supplemental materials.} Further details of the data collection are available in \autoref{sec:app-hit}.

\paragraph{Reference quality}
Beyond our quality control filtering, we analyzed the following: the average Jaccard distance of responses both for workers against themselves and against all of the provided responses for a prompt, the average number of responses provided by workers, and the fatigue factor for each of the prompt datasets. Across each of our datasets the average Jaccard distance between each reference is high (at or near .9 across the board). Therefore, we conclude that there is \textbf{high diversity among the collected references}. This fact is key to the success of evaluation using multiple references~\citep{gangal-etal-2021-improving}. If the references are not diverse, using multiple references is barely better than using one reference. Also, we observed that as a worker completed a HIT, they provided fewer responses per prompt. This is a sign of worker fatigue. Consequently, having longer HITs can decrease the quantity and potentially the quality of collected data (\autoref{fig:ref-jaccard}). 

\subsection{Scraping new conversations}


\href{https://www.rong-chang.com}{\texttt{rong-chang.com}} is a  website that has over 3500 multiturn conversations (10+ turns) on a variety of topics that are used for instructing ESL speakers.
With their explicit permission, we
scrape these conversations from their website and we ask AMT workers to create references for 1000 randomly sampled snippets of 2 or 3 turns.
 Ultimately, we obtain a wide variety of conversation topics and conversations. With this dataset we are consistently able to collect more responses per prompt, which we attribute to the naturalness of the conversations.


\section{Methodology}

To validate the utility of our dataset, we ask a few basic questions about the metrics. In particular, we aim to validate or challenge relationships between well-established metrics.  
Our approach is to evaluate outputs using multiple references rather than a single reference. For multiple models' responses to the same prompts, we use multiple evaluation metrics to score each of them. 

We explore: (1) The Pearson and Spearman correlation between metric evaluations and human evaluations, (2) the Kendall rank correlation coefficient between metric evaluations and human evaluations, and (3) the relationship between output similarity and metric evaluations.

\section{Models}
\label{sec:params}
In order to understand how different metrics are able to distinguish between the quality of different models, we would like a large, diverse collection of model outputs. 
We collect these by using large pretrained models, and by training our own models. 

The pretrained models we use are: Blenderbot \cite{shuster2022blenderbot}, 
Open-Assistant SFT-4 \cite{Open-Assistant},  Koala  \cite{koala_blogpost_2023}, MPT-7B-Chat \cite{MPT_blogpost_2023}, FastChat-T5 \cite{zheng2023judging}, and Vicuna \cite{vicuna2023}. We employ a variety of temperature and sampling strategies for each model. 


Following \citet{khayrallah-sedoc-2020-SMRT}, 
we train Transformer \cite{transformer} chatbots in \textsc{fairseq} using base parameters from the \textsc{flores} benchmark for low-resource MT \cite{guzman-etal-2019-flores}. In order to explore the full space of models with a variety of performance levels, we perform a hyperparameter sweep of regularization parameters, including  SentencePiece  \cite{kudo-richardson-2018-sentencepiece} vocabulary size, dropout,	attention \& relu dropout, and  label smoothing. We also use 8 different decoding strategies. We train on the DailyDialog corpus \cite{li-etal-2017-dailydialog}, as released by ParlAI \cite{miller-etal-2017-parlai}.

For replication and hyper parameter info see \autoref{app:dialog_models}.

\begin{figure*}[tbh]
  \includegraphics[width=1.1\linewidth]{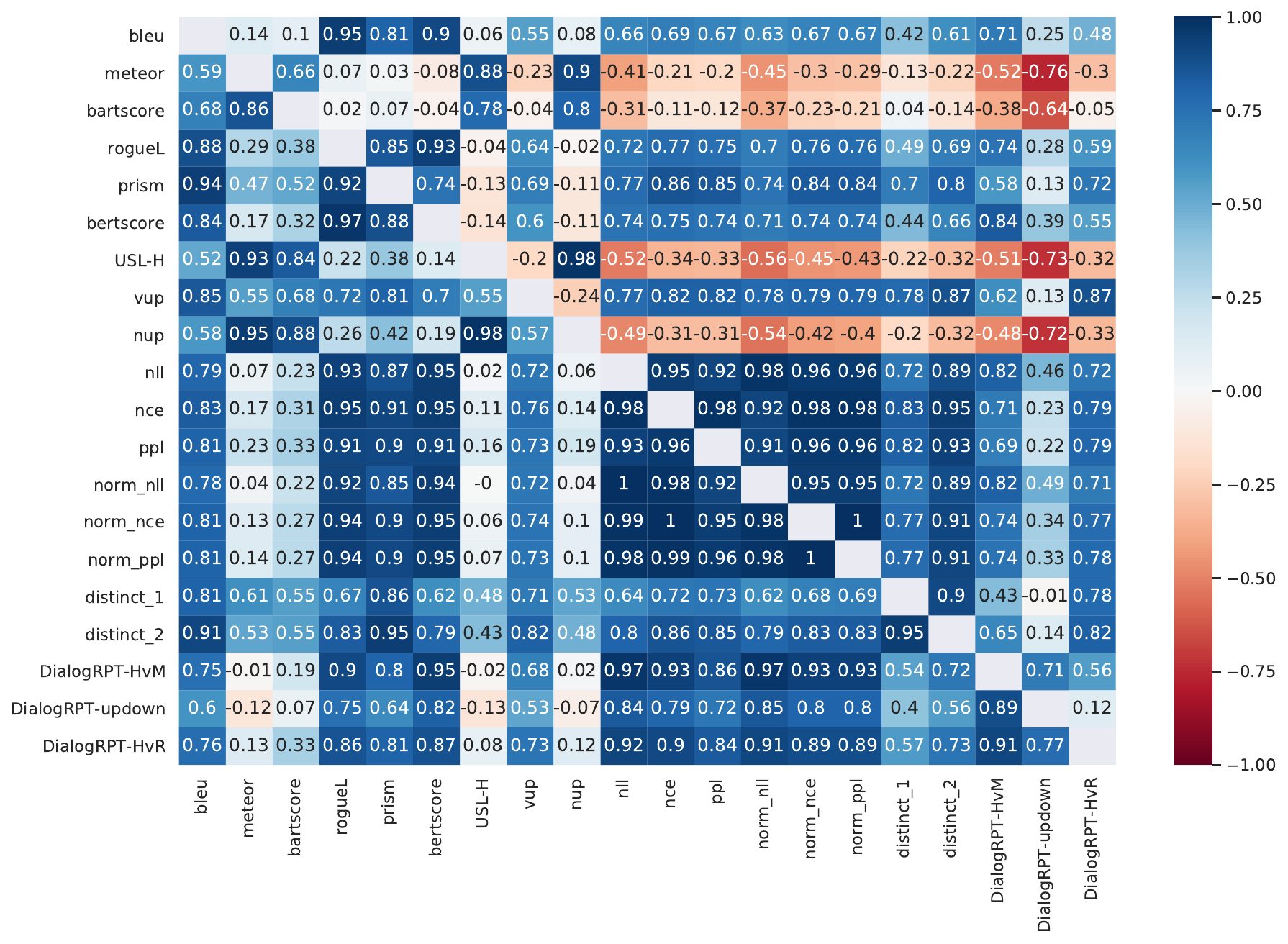}
  \caption{Spearman correlations between various metrics on the ESL3 test set. The bottom left includes all systems, the top right is the top ones.}
  \label{fig:esl3_0.99_spearman}
\end{figure*}

\begin{figure}[tbh]
  \includegraphics[width=\linewidth]{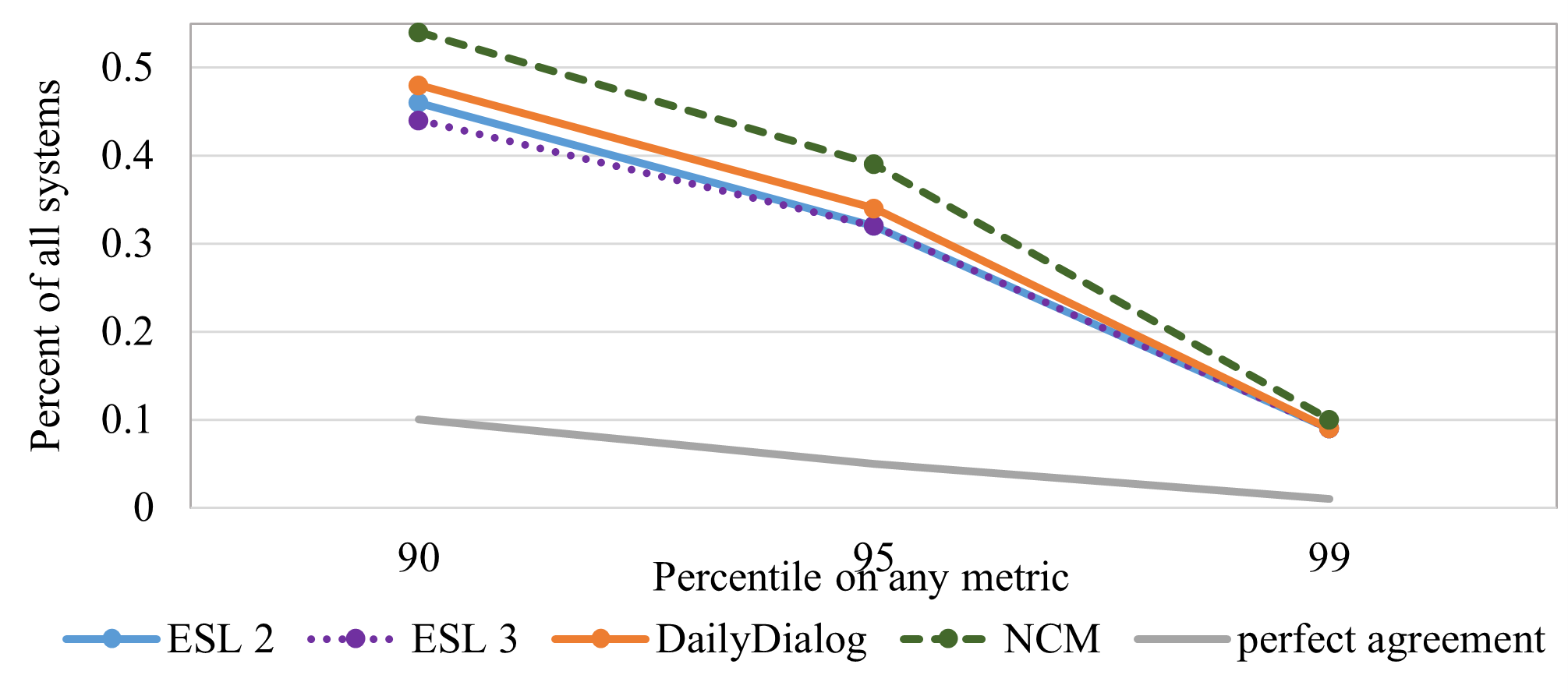}
  \vspace{-20pt}
  \caption{The percent of data retained when thresholding on a percentile for any of the metrics. The dotted grey line shows the percentage that would be retained if all metrics were in perfect agreement.}
  \label{fig:percentilepercents}
  \vspace{-10pt}
\end{figure}

\section{Analysis}

%

\citet{mathur-etal-2020-tangled} showed that correlating a machine translation metric with human judgments is far easier when considering all systems (including very weak ones) than when only considering top systems. Text simplification metrics also have similar behavior, where the correlation between metrics and human judgments decreases when filtered by system quality~\citep{alva-manchego2021unsuitability}.

This is somewhat intuitive: truly terrible systems are easier to differentiate from good ones. Therefore, we consider how well the metrics correlate overall, and when only considering the top systems. 

We define top scoring as any system that is in the 99th percentile of systems on any metric. \autoref{fig:percentilepercents} shows that top scoring systems constitute a large percentage of systems overall, which further highlights the disagreement between metrics. 48\% of the systems are in the 90th percentile or above on some metric for NCM. If the metrics were in perfect agreement, only 10\% of system would be in the 90th percentile.  With so little agreement, it can be particularly hard to know which metrics to trust, highlighting the need for such a dataset for further metrics research. 
  \autoref{fig:esl3_0.99_spearman} shows Spearman correlations between the various metrics (also see additional tables in the appendix). The bottom left half of each table shows the correlation between the metrics on all systems. The top right half shows the correlation between the top scoring systems.


Unsurprisingly, correlations are much stronger overall when comparing all systems rather than only comparing the top systems. 

Meteor, Bartscore, USL-H and nup do not correlate well with other metrics, and have negative correlations in many settings. We note that Metor and Bartscore correlate well with each other. A lack of correlation with other metrics is not necessarily an indication of quality of a particular metric. But rather, this shows that there is poor agreement amongst metrics, and the metric chosen can have a large impact on the final ranking of ssystems. 

\section{Conclusion}

We (will, upon publication) release MMSMR, a \textbf{M}assively \textbf{M}ulti-\textbf{S}ystem \textbf{M}ulti\textbf{R}eference dataset to enable future work on metrics and evaluation for dialog.
The dataset contains 1000 two and three-turn prompts with multiple human-generated references. We train 1750 systems and evaluate them on our novel test set and the DailyDialog dataset. We also evaluate publically available models on these data sets. Our analysis of the metrics shows that the correlations are lower when considering only the top systems than when considering all systems. Our findings show the utility of this novel test set, and  model hyper parameters, inference outputs, and metric scores for each system on a variety of datasets.

\begin{singlespace}
\bibliographystyle{acl_natbib}
\bibliography{anthology,lit}
\end{singlespace}
\clearpage
\appendix

\section*{Appendix}

\section{Dialog Models}
\label{app:dialog_models}

The pretrained models we use are: Blenderbot \cite{shuster2022blenderbot},\footnote{\url{https://huggingface.co/facebook/blenderbot-3B}} 
Open-Assistant SFT-4 \cite{Open-Assistant},\footnote{\url{ https://huggingface.co/OpenAssistant/oasst-sft-4-pythia-12b-epoch-3.5}}  Koala  \cite{koala_blogpost_2023},\footnote{\url{https://huggingface.co/TheBloke/koala-13B-HF}} MPT-7B-Chat \cite{MPT_blogpost_2023}, FastChat-T5 \cite{zheng2023judging},\footnote{\url{https://huggingface.co/lmsys/fastchat-t5-3b-v1.0}} and Vicuna \cite{vicuna2023}.\footnote{\url{https://huggingface.co/lmsys/vicuna-13b-delta-v1.1}} We employ a variety of temperature and sampling strategy for each model. We use temperatures varied from 0.6 to 2.55 in increments of 0.05. We use  top p sampling values of 0.7, 0.8, 0.9, 0.10; top k sampling values of 40, 50, 60, 70, and beam search beams of 1, 2, 3, 4. 






We train Transformer conditional language models in \textsc{fairseq} using parameters from the \textsc{flores}\footnote{\url{https://github.com/facebookresearch/flores/tree/5696dd4ef07e29977d5690d2539513a4ef2fe7f0}} benchmark for low-resource machine translation \cite{guzman-etal-2019-flores}.

As a baseline, we use a  $5$-layer encoder and decoder, $512$ dimensional embeddings, and $2$ encoder and decoder attention heads. We regularize with $0.2$ label smoothing, and $0.4$ dropout. 
We optimize using Adam with a learning rate of $10^{-3}$. 
We train 100 epochs, and select the best checkpoint based on validation set perplexity. We run inference several ways: greedy search,
beam size 10, beam size 100, top p=.5 sampling, top p=.7 sampling, top p=.9 sampling, top k=10, top k=100. We do not use a length penalty.

We sweep SentencePiece  \cite{kudo-richardson-2018-sentencepiece} vocabulary size (1k,2k, 4k,8k,16k),  dropout (0.0, 0.1, 0.2, 0.3, 0.4),	attention \& ReLU dropout (0.0, 0.1, 0.2, 0.3, 0.4), and  label smoothing (0.0, 0.1, 0.2, 0.3, 0.4, 0.5, 0.6).

%
 \autoref{fig:nll_train} shows the train command.

\begin{figure*}[h]
\begin{verbatim}
python train.py \
 $DATADIR \
 --source-lang src \
 --target-lang tgt \
 --seed 10 \
 --save-dir $SAVEDIR \
 --patience 50 --criterion label_smoothed_cross_entropy \
 --label-smoothing 0.2 \
 --share-all-embeddings \
 --arch transformer  --encoder-layers 5 --decoder-layers 5 \
 --encoder-embed-dim 512 --decoder-embed-dim 512 \
 --encoder-ffn-embed-dim 2048 --decoder-ffn-embed-dim 2048 \
 --encoder-attention-heads 2 --decoder-attention-heads 2 \
 --encoder-normalize-before --decoder-normalize-before \
 --dropout 0.4 --attention-dropout 0.2 --relu-dropout 0.2 \
 --weight-decay 0.0001 \
 --optimizer adam --adam-betas '(0.9, 0.98)' --clip-norm 0 \
 --lr-scheduler inverse_sqrt --warmup-updates 4000 --warmup-init-lr 1e-7 \
 --lr 1e-3 --min-lr 1e-9 --no-epoch-checkpoints \
 --max-tokens 4000 \
 --max-epoch 100 --save-interval 10 --update-freq 4 \
 --log-format json --log-interval 100 
\end{verbatim}
\caption{Training command.}
\label{fig:nll_train}
\end{figure*}

We evaluate on the DailyDialog corpus \cite{li-etal-2017-dailydialog}, as released by ParlAI \cite{miller-etal-2017-parlai}.\footnote{\url{https://github.com/facebookresearch/ParlAI/tree/1e905fec8ef4876a07305f19c3bbae633e8b33af}} 
We train both a single and multiturn model. We evalute DailyDialog and NCM on the single turn models, and ESL2/3 on the multiturn models.

\section{Human Annotation Details}
\label{sec:app-hit}
For the NCM dataset 3 workers responded to each conversation, and for every other dataset, 4 workers responded to each conversation. Workers were informed that they would receive an extra cent as bonus for each response provided beyond the minimum required two per conversation. The task itself paid thirty cents, which we now realize was too low for the difficulty and time requirement. 
The maximum a worker could receive was sixty cents(for providing every `extra' response, thirty cents for the HIT and thirty cents in bonus). A quality control check was not included in the HIT itself but was performed after results were collected and before approving or rejecting assignments. We filtered out and rejected workers who provided responses that either were: not unique, one character, or punctuation only. This constituted a small fraction of workers. 

\subsection{Dissimilarity of References}
For every conversation in each of the datasets we have anywhere from 6-20 responses. We noticed an inverse relationship between the prompt number and the average number of responses from workers. 

\begin{figure}
    \centering
    \includegraphics[width=0.9\linewidth]{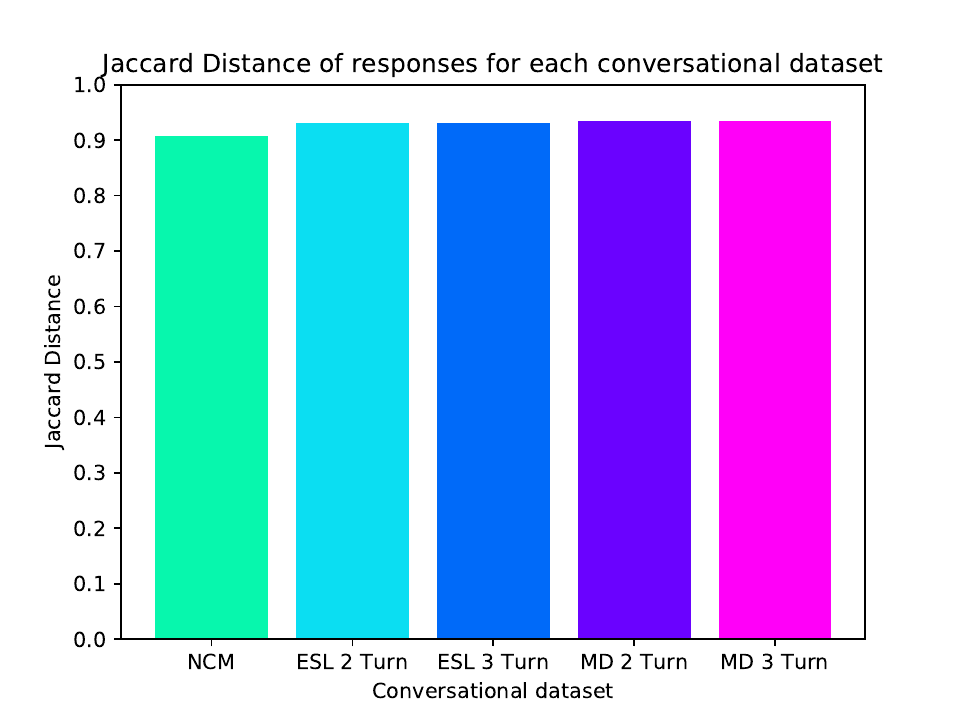}
    \caption{Jaccard index of multiple references}
    \label{fig:ref-jaccard}
\end{figure}

	Using the Jaccard distance as for quantifying diversity in responses, we found that the ESL dataset had the greatest diversity. 
	However, even single turn prompts from the NCM got diverse responses. For example, the prompt "What is two plus two?" from the NCM dataset got responses such as: "four", "same as five plus three", and "I'm 3, how would I know?" with each of these answers coming from a different worker. \autoref{fig:ref-jaccard} shows the Jaccard distance scores for each of the datasets.
 
\end{document}